# From Narratives to Probabilistic Reasoning: Predicting and Interpreting Drivers' Hazardous Actions in Crashes Using Large Language Model

Boyou Chen, Ph.D. Student
Industrial and Manufacturing Systems Engineering Department
University of Michigan-Dearborn, 4901 Evergreen Rd, Dearborn, MI, USA
Email: boyou@umich.edu

Gerui Xu, Ph.D. Student
Industrial and Manufacturing Systems Engineering Department
University of Michigan-Dearborn, 4901 Evergreen Rd, Dearborn, MI, USA
Email: geruix@umich.edu

Zifei Wang, Ph.D. Student
Industrial and Manufacturing Systems Engineering Department
University of Michigan-Dearborn, 4901 Evergreen Rd, Dearborn, MI, USA
Email: zifwang@umich.edu

Huizhong Guo, Ph.D.
University of Michigan Transportation Research Institute, 2901 Baxter Road, Ann Arbor, MI, 48109, USA
Email: hzhguo@umich.edu

Ananna Ahmed, Ph.D.
Toyota Motor Engineering & Manufacturing North America, Inc., USA
Email: ananna.ahmed@toyota.com

Zhaonan Sun, Ph.D.
Toyota Motor Engineering & Manufacturing North America, Inc., USA
Email: zhaonan.sun@toyota.com

Zhen Hu, Ph.D.
Industrial and Manufacturing Systems Engineering Department
University of Michigan-Dearborn, 4901 Evergreen Rd, Dearborn, MI, USA
Email: zhennhu@umich.edu

Kaihan Zhang, Ph.D. Student
Cho Chun Shik Graduate School of Mobility
Korea Advanced Institute of Science and Technology
193 Munji–ro, Yuseong–gu, Daejeon, 34051, South Korea
Email: kaihn@kaist.ac.kr

Shan Bao, Ph.D.* Corresponding Author
Industrial and Manufacturing Systems Engineering Department

University of Michigan-Dearborn, 4901 Evergreen Rd, Dearborn, MI, USA
University of Michigan Transportation Research Institute, 2901 Baxter Road, Ann Arbor, MI, 48109, USA
Email: shanbao@umich.edu

# ABSTRACT

Vehicle crashes involve complex interactions between road users, split-second decisions, and challenging environmental conditions. Among these, two-vehicle crashes are the most prevalent. Accounting for approximately 70% of roadway crashes, they present a significant challenge to traffic safety. Identifying Driver Hazardous Action (DHA) is essential for understanding crash causation, yet the reliability of DHA data in large-scale databases is limited by inconsistent and labor-intensive manual coding practices. Here, we present an innovative framework that leverages a fine-tuned large language model (LLM) to automatically infer DHAs from textual crash narratives, thereby improving the validity and interpretability of DHA classifications. Using five years (2019–2023) of two-vehicle crash data from Michigan Traffic Crash Facts (MTCF), we fine-tuned the Llama 3.2 1B model on detailed crash narratives and benchmarked its performance against conventional machine learning classifiers, including Random Forest, XGBoost, CatBoost, and a neural network. The fine-tuned LLM achieved an overall accuracy of 80%, surpassing all baseline models and demonstrating pronounced improvements in scenarios with imbalanced data. To increase interpretability, we developed a novel probabilistic reasoning approach, analyzing model output shifts across original test sets and three targeted counterfactual scenarios: variations in driver distraction and age. Our analysis revealed that introducing distraction for one driver substantially increased the likelihood of "General Unsafe Driving"; distraction for both drivers maximized the probability of "Both Drivers Took Hazardous Actions"; and assigning a teen driver markedly elevated the probability of "Speed and Stopping Violations." Together, our framework and analytical methods provide a robust and interpretable solution for large-scale automated DHA detection, offering new opportunities for traffic safety analysis and intervention.

# 1. INTRODUCTION

Vehicle crashes are complex events arising from the interplay of challenging environmental conditions, the need for split-second decisions, and intricate interactions among road users [1, 2, 3]. This complexity places significant cognitive load and time pressure on drivers, who must continuously perceive information, process risks, and execute maneuvers [4, 5, 6]. Given that studies consistently attribute the majority of crashes to human error [7, 8, 9, 10], identifying leading driver behaviors prior to crashes is critical for effective crash causation analysis. This is especially pertinent for two-vehicle crashes, such as rear-end and angle collisions, which represent the most prevalent crash type, accounting for a substantial majority of all reported roadway incidents [8]. The high frequency of these events underscores their significant burden on public health and safety systems, making the understanding of pre-crash behavior a central focus for traffic safety research and policy intervention. On Michigan's UD-10 crash report, such actions are captured in the Driver Hazardous Action (DHA) field, where the investigating officer records the single maneuver judged most responsible for precipitating the crash [11]. Because this field succinctly reflects the officer's opinion of primary driver fault, it serves both as a proxy for liability and as a valuable starting point for safety analysis. Therefore, systematically identifying DHAs and further understanding the reasons behind them fulfils two complementary roles. First, it aids crash reconstruction and legal deliberations by pinpointing the behavior most likely to have caused the crash [11]. Second, aggregating DHA patterns across contexts (for instance, pedestrian crash and vehicle crash) highlights the risky maneuvers that disproportionately contribute to crashes, thereby guiding targeted countermeasures [12, 13]. Furthermore, enriching national crash databases such as the Crash Report Sampling System (CRSS) and the Fatality Analysis Reporting System (FARS) with a dedicated DHA variable would be advantageous, as although both systems document multiple driver-related factors, neither identifies the single most significant [8]. However, accurate DHA coding is challenging and labor-intensive, as officers must weigh multiple, sometimes interacting factors under time pressure, and uncommon crash configurations can further cloud judgment. While the DHA field offers a concise insight through which to study driver error in crashes, its reliability depends on consistent, high-quality coding, resulting in an issue that needs to be addressed to advance research in crash prevention.

Recent large language models (LLMs) can process tasks presented in diverse formats and return sophisticated, human-like answers. Emerging research has demonstrated the effectiveness and scientific validity of adopting LLM in traffic crash analysis [14, 15]. Unlike conventional crash analyses through classification-based approaches, such as machine learning (ML) and statistical models that rely solely on the data explicitly provided to them at training and inference time. In contrast, LLMs not only learn from the given dataset but also incorporate background information during pre-training, even when it's not directly present in the input. Previous studies have applied natural language processing (NLP) and ML techniques to identify DHAs from police crash narratives; however, they have been limited to narrowly scoped tasks, typically predicting a binary outcome (hazardous vs. non-hazardous) or a small subset of DHA categories, and have relied on selected samples due to limited narrative quality [13, 16]. A scalable framework and approach that can recover the complete spectrum of DHA codes across large crash databases remains an unmet need. Overall, although LLMs show considerable promise for crash-analysis tasks, their internal reasoning remains opaque, creating a "black-box" barrier that reduces confidence among crash analysts, law-enforcement personnel, and policymakers. Therefore, a tool for identifying DHAs should incorporate an explainable approach that makes LLM's outputs interpretable [17, 18].

In this study, we develop and validate an LLM-based framework for automatically inferring the primary DHA in two-vehicle crashes, utilizing data from Michigan Traffic Crash Facts (MTCF). First, we fine-tune the Llama 3.2 1B model on MTCF data for DHA classification and benchmark its performance against four established baselines, Random Forest, XGBoost, CatBoost, and a feed-forward neural network, to ensure robust and comparative evaluation. Second, we introduce a novel probabilistic reasoning approach that simulates counterfactual scenarios by varying drivers' distraction statuses and age attributes. By tracking changes in the predicted probability distribution across DHA classes, we quantify how the model's probability output shifts in response to unobserved but traditionally high-risk crash factors.

This study contributes to the emerging literature on LLM-enabled traffic-safety analytics in three ways. First, it offers the first large-scale, multi-class DHA classifier that matches officer-coded labels while operating on reformulated crash narratives derived from tabulated data, providing a scalable alternative where DHA is missing or inconsistently coded. Second, the proposed probability-shift analysis furnishes an interpretable lens on the model's decision process, transforming a black-box predictor into a transparent decision-support tool for analysts and policymakers, as demonstrated in the two-vehicle scope. Third, by demonstrating reasonable performance on Michigan data and outlining a straightforward mapping between UD-10 DHA codes and the driver-factor taxonomies used in CRSS and FARS, the study lays the groundwork for extending automated DHA coding nationwide. Collectively, these advances move the field toward richer, more explainable crash-causation insights that can underpin targeted safety interventions.

## 2. LITERATURE REVIEW

**2.1. Driver Hazardous Action (DHA)**

As mentioned above, DHAs represent the maneuvers that investigating officers judge to have precipitated a crash. Because the DHA field distills unstructured evidence into a single, officer-interpreted fault code for each driver, it is indispensable for liability assessment and for pinpointing behaviors that systematically elevate crash risk. Existing studies in the MTCF corpus demonstrated the practical value of these codes. In a statewide commercial-vehicle study, "failure to yield", "improper lane use", "speed too much", and "unable to stop in assured clear distance" emerged as the most frequent contributory actions in truck-related crashes [19]. Convolutional neural-net (CNN) image analysis was fused with NLP to enumerate risk factors in horse-and-buggy crashes; "careless driving", "poor nighttime visibility", and "failure to yield" were the dominant hazards, especially among older motorists [20]. A companion investigation of micro-mobility crashes found that motor-vehicle left turns and rider lane violations often coupled with driver distraction, markedly increased injury severity [21]. For older drivers at stop-sign-controlled intersections, turning and improper lane changes sharply raised the odds of being deemed at-fault (DHA presented), whereas simply proceeding straight reduced liability. Overall, these insights confirm the practical value of the DHA field [10]. Yet coverage is uneven: certain codes are scarce in some demographic groups and crash types, limiting our understanding of why those maneuvers arise in high-risk contexts. Several studies have also tried to identify DHAs through NLP. Zhang et al. [13] utilized NLP on 276 randomly selected crash narratives in the MTCF database, combined uni-, bi-, and tri-grams to automate the recoded binary DHA classes (whether DHA was present or not) and reported 92.8% accuracy on held-out data. Kwayu et al. [16] extended the approach and mined crash narratives at signalized intersections and, with a support vector machine model using mixed uni- and bi-gram features, achieved 86% out-of-sample accuracy in distinguishing "fail to yield" from "disregard traffic control" DHAs. Because MTCF crash narratives vary widely in detail and structure, identifying DHAs from them is challenging, particularly across diverse crash scenarios. Thus, although NLP models performed well on narrowly defined tasks, a scalable method that can classify the full DHA taxonomy remains limited.

**2.2. LLM Potentials on Crash Analysis**

LLM has demonstrated strong potential in crash-related prediction tasks [22]. A BERT model trained on approximately 750,000 crash narratives classified injury severity with 84% accuracy and an AUC of 0.93 [23], while a cross-cultural BERT + Bi-LSTM ensemble achieved about 99% accuracy on matched U.S. and Jordan datasets, showing strong transferability across dialects and road contexts [24]. Domain-tuned LLMs can also tackle class imbalance. For example, a RoBERTa model was fine-tuned to label pedestrian-crash typology on nearly 30,000 Texas reports, and demonstrated improved minority-class performance [25]. ChatGPT, Bard, and GPT-4 correctly answered work-zone and other binary questions in 70–96% of 100 Iowa/Kansas crash narratives but stumbled on multi-step causal queries [26]; adding chain-of-thought prompts with GPT-3.5-turbo and LLaMA-3 (8 B/70 B) raised zero-shot crash-severity prediction accuracy by up to 9% points [27]. The CrashSage pipeline illustrated end-to-end gains: it converted tabular crash data to enriched narratives, fine-tuned LLaMA-3-8B and coupled gradient explanations, outperforming zero-shot GPT-4o and conventional tabular transformers on Georgia data [18]. Beyond prediction, fine-tuned GPT-3.5 models now audit data quality, flagging mislabeled work-zone crashes and markedly reducing manual review effort [28]. As studies adopting LLM in various crash causation analyses continue to rapidly emerge, DHA identification can also be beneficial from this advanced technique.

## 2.3. Explainable LLM Output

Given that LLMs are often "black boxes," several studies emphasized making their outputs interpretable in the traffic safety context. One strategy is to have the model produce human-readable reasoning (as in the chain-of-thought approach) to justify its conclusions [22, 27]. Another approach is post-hoc explanation: Jaradat et al. [24] used Shapley Additive Explanations (SHAP) with a BERT-based classifier to identify which words in crash narratives most influenced the model's severity predictions, similar SHAP adoption was found in [15], who demonstrated the potential of SHAP in explaining crash contributing factors. Zhen and Yang [18] also utilized gradient-based saliency maps to increase transparency by pinpointing factors driving each prediction. However, a notable gap remains in ensuring that such explanations are faithful to the model's actual decision process. As recent analyses caution, LLM-generated rationales might sound plausible without truly reflecting the underlying computation [29]. Verifying the correctness of model explanations (and not just the correctness of predictions) is an ongoing challenge.

# 3. METHODOLOGY

## 3.1. Analytical Framework

To demonstrate our end-to-end analytical framework for reasoning about DHAs in two–vehicle crashes, the overview of the workflow is proposed in Figure 1. At its core, the pipeline integrates (1) data preprocessing, (2) LLM–based textual reasoning, and (3) probability-shift analysis under counterfactual scenarios. First, raw crash records from the MTCF database were filtered to include two-vehicle crashes with selected pre-crash variables and a recoded DHA group. Structured fields, including covering pre-crash context (crash information, driver information, road characteristics, and environment information) were concatenated with each narrative to form a single input sequence (details see Section 3.2 & 3.3).

Next, these reformulated crash narratives served as prompts to a fine-tuned Llama 3.2 1B model via a causal next-token generation paradigm (details see Section 3.4). Finally, a novel probability reasoning approach was proposed to investigate three representative risky crash scenarios beyond those sampled in the empirical record, we introduced systematic perturbations to the testing set, and for each scenario, the model's output logits are converted to a standardized probability distribution over the seven DHA targets. Two complementary distribution-shift metrics were further computed to quantify how the model's predicted probability reallocates under each counterfactual setting (details see Section 3.5).

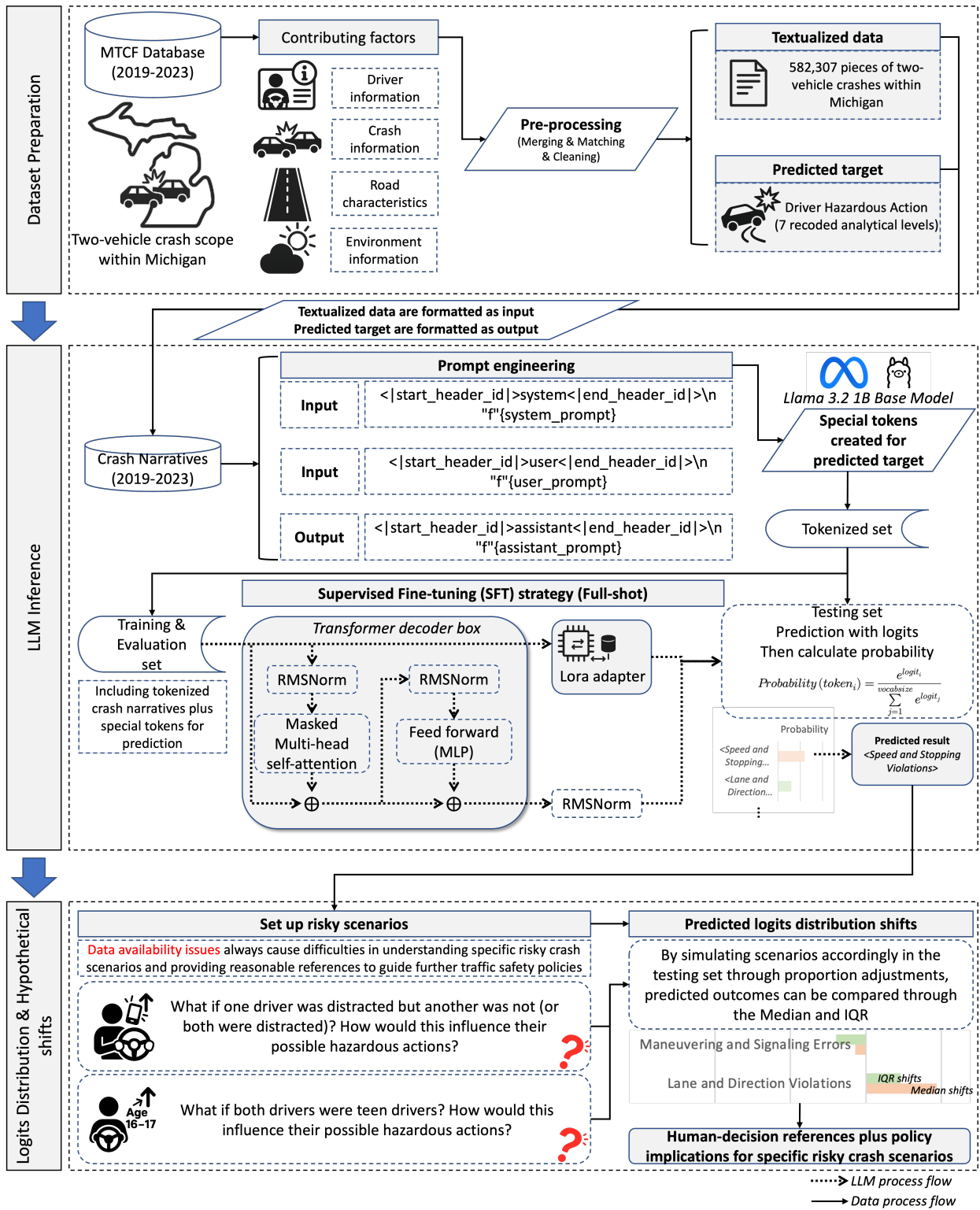

*Figure 1 Overview of the Proposed Analytical Framework.*

**3.2. Data Source**

Crash data were obtained from the MTCF database. The retrieval covered the five most recently completed calendar years available at the time of analysis (January 1, 2019 –December 31, 2023). Records were narrowed down to crashes that (1) involved exactly two motor vehicles, (2) only driver-related records were retained, other person types were excluded, and (3) carried a valid coded DHA. "Other" and "Unknown" in DHA were removed according to UD-10 Traffic Crash Report Instruction Manual [11]; thus, the rest 14 valid levels in DHA were used for analytical purposes. As a result, the initial set included 1,620,138 two-vehicle crash records from 2019 to 2023 (after removing 282,573 records with invalid DHA codes).

**3.3. Potential Contributing Variables**

We identified 18 potential DHA contributing independent variables reflecting pre-crash situation and organized them into four conceptual categories: Crash Information ($n = 6$), Driver Information ($n = 5$), Road Characteristics ($n = 5$), and Environmental Conditions ($n = 2$). Variable selection was guided by a systematic review of recent DHA-related studies and a meticulous manual appraisal of each field in the MTCF database [3, 10, 21]. Variables capturing abnormal driver states such as fatigue, sleep, drug use, or alcohol impairment were intentionally excluded to reduce potential model bias, as they constituted fewer than 1% of records and were widely recognized as underreported [19]. Entries containing the value "invalid" were excluded from the dataset. Variables labeled as "uncoded," "unknown," or "other" were kept only if they comprised more than 5% of the observations; otherwise, they were removed. To facilitate robust modelling and interpretability, the 14 officer-assigned valid HA classes in the MTCF database were regrouped into seven higher-order, behaviorally coherent groups (see Table 1). The final set included 1,164,614 two-vehicle crash records for modeling purposes, and Figure 2 illustrates all selected variable level descriptions.

*Table 1 Recoded DHA Descriptions.*

| Recoded Level of DHA | Original Level of DHA | Counts | Proportion |
|---|---|---|---|
| Speed and Stopping Violations (SSV) | Unable to Stop in Assured Clear Distance | 216,022 | 18.55% |
| | Speed Too Fast | 25,286 | 2.17% |
| | Speed Too Slow | 339 | 0.03% |
| Right-of-Way and Traffic Control Violations (RWTCV) | Failed to Yield | 183,069 | 15.72% |
| | Disregard Traffic Control | 39,007 | 3.35% |
| Lane and Direction Violations (LDV) | Improper Lane Use | 43,250 | 3.71% |
| | Drove Left of Center | 6,369 | 0.55% |
| | Improper Passing | 9,676 | 0.83% |
| | Drove Wrong Way | 1,106 | 0.10% |
| Maneuvering and Signaling Errors (MSE) | Improper Turn | 19,763 | 1.70% |
| | Improper Backing | 15,651 | 1.34% |
| | Improper/No Signal | 1,705 | 0.15% |
| General Unsafe Driving (GUD) | Careless/Negligent Driving | 14,846 | 1.28% |
| | Reckless Driving | 3,183 | 0.27% |
| None | None | 585,342 | 50.26% |

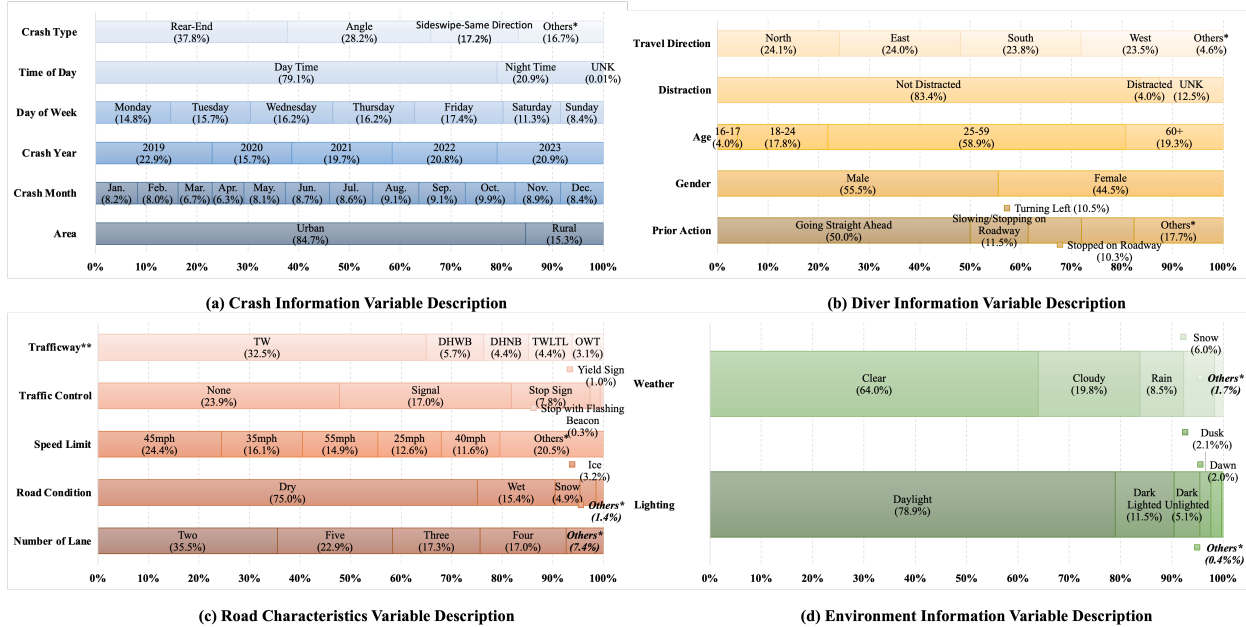

*Figure 2 Independent Variable Level Descriptions.* (a) illustrates two-vehicle crash-related information, (b) illustrates involved driver-related information, (c) illustrates on-site road characteristics, and (d) illustrates on-site environmental information. All proportions were calculated based on a total sample size of 1,164,614 two-vehicle crash records. Notably, "***Others\****" in "Crash Type", "Travel Direction", "Speed Limit", "Road Condition", "Number of Lane", "Weather", "Lighting" indicate other levels (including unknown) that account for a small proportion, but kept original levels in the modelling process; TW, DHWB, DHNB, TWLTL, and OWT in "***Trafficway\*\****" refer to two-way traffic (not physically divided), divided highway with traffic barrier, divided highway without traffic barrier, two-way with continuous left-turn lane, and one-way, respectively. "Age" kept the original continuous format in the modelling process.

## 3.4. LLM Fine-Tuning for Textualized Crash Reasoning

To leverage the rich contextual cues encoded in free-text crash narratives, we reformulated/assembled 582,307 two-vehicle crash narratives from the 1,164,614 eligible records described in Section 3.3, into a next-token-generation task. Each narrative was concatenated with its associated structured variables and passed as input to a causal language model based on the Llama 3.2 1B architecture. It should be noted that during the reformulation of two-vehicle crash narratives, three conditions were possible: (1) only one driver performed an HA; (2) both drivers performed HAs; or (3) neither driver performed an HA. To account for these possibilities, we supplemented the four recoded DHA classes with two additional levels, "Both Drivers Took HAs" and "No HAs", which represented 0.264 % and 0.364 % of 582,307 narratives, respectively. As illustrated in Figure 3, each reformulated crash narrative was structured as a three-stage prompt, including system, user, and assistant as this role-based design leveraged the LLM's natural language understanding and to perform the next-token-generation task in a fully contextualized and prompt-driven manner.

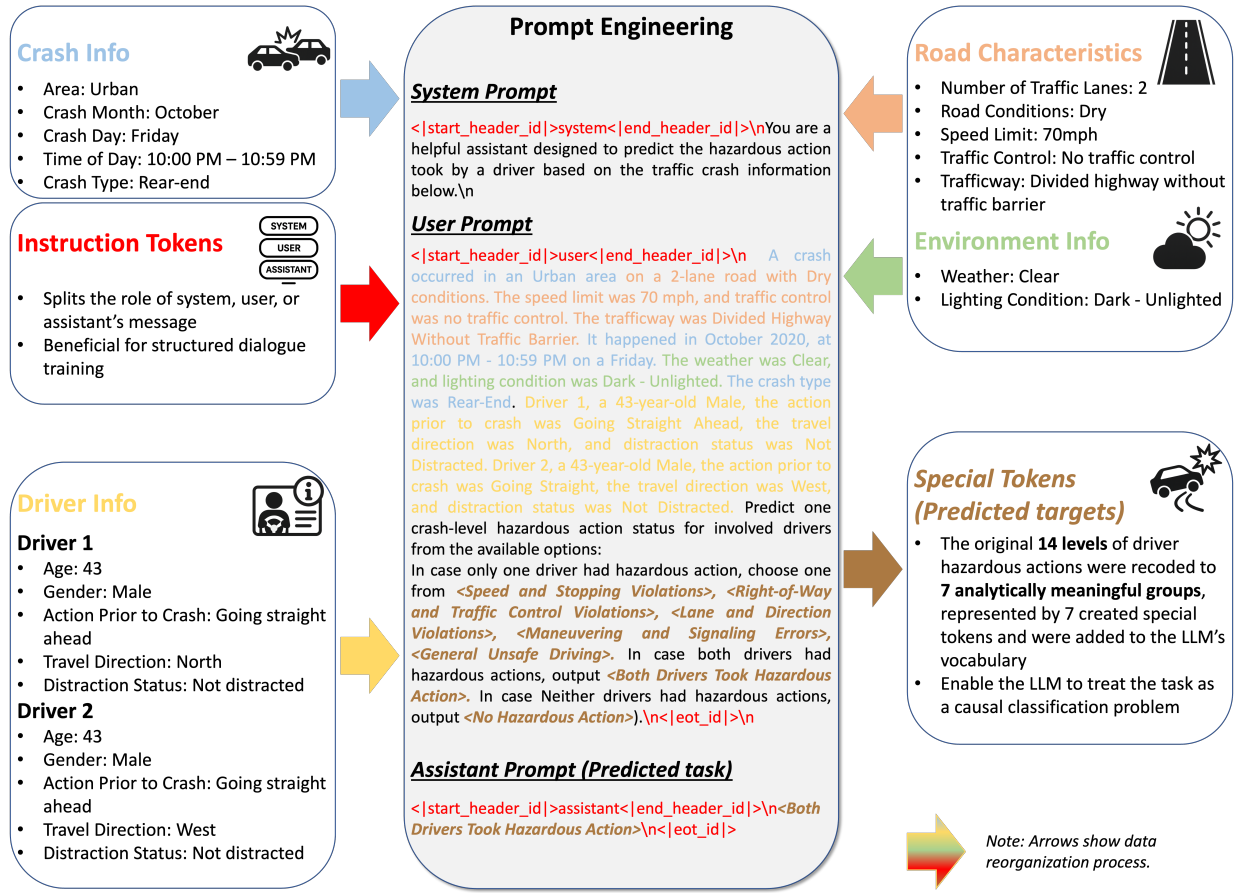

*Figure 3 Illustration of Prompt Design.*

The prompt formulation shown in Figure 3 represents how we concatenated two crash records through the same crash ID into one two-vehicle crash narrative, and how we formulated the narrative into structured, contextualized input and output. The 582,307 crash narratives were randomly split into a training set (70%), evaluation set (15%), and testing set (15%) for fine-tuning purposes. Since the LLM treats all input words as processable tokens based on its vocabulary, after the tokenization process through the Llama 3 tokenizer, a 4-step-fine-tuning process can be described as:

- *Input embedding and positional encoding*

$$\mathbf{h}_i^{(0)} = \mathbf{e}(x_i) + \mathbf{p}(i), \tag{1}$$

where each token $x_i$ ($i$ denoted the index of the token) was mapped to a $d$-dimensional (which was 2048 in our fine-tuned model) embedding $\mathbf{e}(x_i)$, and a fixed positional vector $\mathbf{p}(i)$ was added so that the model can recognize each token's place in the sequence.

- *Transformer decoder block*

$$\begin{cases} \mathbf{a}_i^{(l)} = \text{RMSNorm}\left(\mathbf{h}_i^{l-1} + \text{MHSA}(\mathbf{H}^{(l-1)})_i\right) \\ \mathbf{h}_i^{(l)} = \text{RMSNorm}\left(\mathbf{a}_i^{(l)} + \text{FFN}(\mathbf{a}_i^{(l)})\right) \end{cases}, \quad (2)$$

where for $l = 1, \ldots, L$ ($L = 16$ in our fine-tuned model). The full sequence of embeddings $\mathbf{H}^{(l-1)}$ was fed into a multi-head self-attention (MHSA) layer, then a feed-forward network (FFN), each wrapped with residual connections and layer normalization. Stacking $L$ such blocks built deep contextual representations.

- *Linear projection to logits*

$$\mathbf{z} = W\mathbf{h}_n^{(L)} + \mathbf{b}, \mathbf{z} \in \mathbb{R}^7, \quad (3)$$

where we took the final hidden state at the sequence's end token $\mathbf{h}_n^{(L=16)}$, projected it onto the modified vocabulary of 128,263 tokens, and applied a learned weight matrix $W$ plus bias $\mathbf{b}$ to produce seven unnormalized logits, one per DHA class.

- *Softmax normalization*

$$P(y = k|X) = \frac{\exp(z_k)}{\sum_{j=1}^{7} \exp(z_j)}, k = 1, \ldots, 7. \quad (4)$$

Finally, a softmax converted those logits ($z_k$) into a valid probability distribution over the seven DHA targets. During fine-tuning, we applied LoRA (26) to all projection matrices in the self-attention and feed-forward sublayers, injecting trainable rank-$r$ adapters while keeping the 16-layer Llama 3.2 1B base weights frozen. The model was loaded in 16-bit floating-point precision (FP16) and trained under a mixed-precision regime to maximize memory and compute efficiency.

Only the LoRA adapter parameters were updated, and no changes were made to the original transformer weights, resulting in a > 95 % reduction in tunable parameters. We optimized a cross-entropy loss computed solely over the seven specially-added DHA target tokens, thereby directing the model to maximize the likelihood of generating the correct classification token immediately after each prompt. The entire training process was set up on the University of Michigan Great Lakes Cluster Platform with one A100 40GB GPU. For benchmarking, we compared our fine-tuned LLM against four conventional classifiers, Random Forest, XGBoost, CatBoost and a feed-forward Neural Network, which were trained and evaluated on TF-IDF (term frequency-inverse document frequency) vectorizations of the same 582,307 narratives. Each baseline model received the seven target labels as training outputs.

### 3.5. Probability-Based Situational Analysis

Identifying DHAs typically relies heavily on empirical observations, yet inherently faces limitations when investigating effects of high-risk or infrequently observed scenarios such as those involving multiple distracted drivers or exclusively teen drivers that are poorly represented in existing crash records. Traditional classification-based machine learning models struggle to meaningfully quantify DHA probability shifts under such hypothetical circumstances due to limited data samples and rigid input-output frameworks. To address these analytical constraints, this study adopts a probability-based reasoning approach, leveraging the human-like inference capabilities of a fine-tuned LLM. By systematically perturbing the testing set with synthetic scenario modifications, specifically, single-driver and both-driver distraction labelling and substituting teenage drivers (aged 16–17), our model quantifies how these hypothetical crash scenarios affect the predicted probabilities of DHAs. A summary of the three hypothetical risky crash scenarios' setup is provided below:

- **Single-driver Distraction Scenario**: This scenario synthetically modified each crash narrative in the testing set by relabeling exactly one random driver as "distracted" and the other as "not distracted." Originally, distracted drivers accounted for approximately 4% of the cases; this modification effectively increased the distraction presence to 50% across involved drivers, enabling exploration of single-driver distraction effects on DHAs.

- **Both-driver Distraction Scenario**: Building on the previous scenario, this condition further intensified the risk by labeling both drivers as "distracted" in every crash narrative, thus increasing the proportion of distracted drivers from the original 4% to a full 100%. This scenario allowed examination of model predictions under the most extreme distraction circumstances, providing insights into how the presence of Both-driver distraction influences the model's predicted probability and DHAs outcomes.

- **Teen Driver Scenario (Age 16–17)**: This scenario replaced all original drivers with teen drivers (aged either 16 or 17), thereby increasing their proportion from approximately 4% in the baseline to 100%. Teen drivers were specifically chosen because existing empirical literature consistently highlights their distinct and elevated crash risks. This scenario was designed to evaluate how the model reallocates probability among DHAs when confronted exclusively with teen driver demographic.

After getting predicted probability for each crash narrative in the testing set (serves as the baseline situation), and each in three hypothetical crash scenarios. Two analytical indicators ($\Delta med$, refers to median predicted probability change proportion; $\Delta IQR$, refers to probability interquartile range change proportion; calculation follows **Equations 5 & 6**) were used to evaluate how the fine-tuned LLM's predicted probability distribution shifts based on comparisons between the baseline and three proposed hypothetical risky crash scenarios.

$$\Delta_{med} = \frac{P_{med}^{scenario} - P_{med}^{baseline}}{P_{med}^{baseline}} \times 100\%, \tag{5}$$

$$\Delta_{IQR} = \frac{IQR_{scenario} - IQR_{baseline}}{IQR_{baseline}} \times 100\%, IQR = Q_3 - Q_1, \tag{6}$$

where $P_{med}^{scenario}$ and $P_{med}^{baseline}$ refer to LLM's predicted probability median values across all crash narratives in the hypothetical scenario and the baseline (the original testing set). $IQR_{scenario}$ and $IQR_{baseline}$ denote LLM's predicted probability interquartile range (75% quantile probability $Q_3$ minus 25% quantile probability $Q_1$) in the hypothetical scenario and the baseline. We chose not to compare raw class-level probabilities across scenarios because the model was trained on highly imbalanced data: rarer classes typically receive smaller probability mass simply to satisfy the unit-sum constraint. Instead, we examined how the entire predicted distribution shifts, which more meaningfully reveals changes in the model's predicted probability without being dominated by base-rate effects. This approach uniquely allows us to explore these risky scenarios beyond the empirical data limitations, revealing reasonable shifts in model predictions and generating critical insights that can inform traffic safety decisions.

## 4. RESULTS

### 4.1. Model Performance Evaluation

As shown in Table 2 and illustrated by the confusion matrix comparisons in Figure 4, the fine-tuned Llama 3.2 1B model substantially outperformed all baseline classifiers on the two most prevalent DHA classes ("SSV" and "RWTCV"), achieving F1 scores of 0.91 and 0.83, respectively, and yielding the highest overall accuracy of 0.80. Its weighted F1 of 0.77 also exceeded that of Random Forest (0.70), XGBoost (0.69), CatBoost (0.67), and the Neural Network (0.51). In contrast, performance on low-support classes ("LDV", "MSE", "GUD", "NHA", "BDTHA") remained challenging across all models, in which the fine-tuned LLM's Macro-average F1 of 0.46 indicated considerable confusion among rarer classes, particularly "GUD," "NHA," and "BDTHA," which were frequently misassigned to dominant classes. However, these limitations also affected baseline models, and the fine-tuned LLM still outperformed these benchmarks with the highest macro F1 score of 0.46.

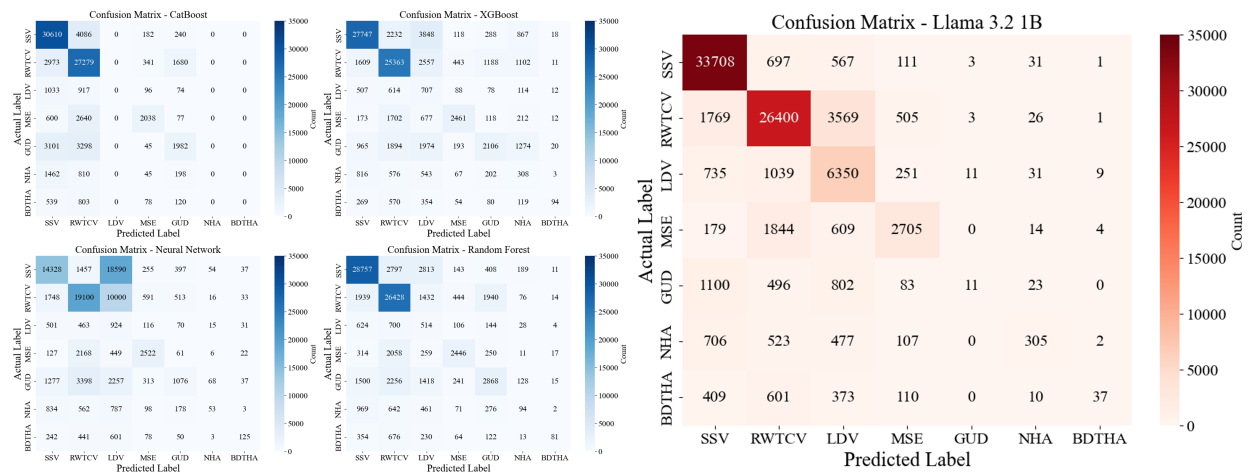

*Figure 4 Model Confusion Matrix Comparisons.* All confusion matrices were performed based on the testing set (87,347 samples, 15%) among the five proposed models. Five baseline models were trained with grid-search optimizations. Seven classification targets: "SSV", "RWTCV",

"LDV", "MSE", "GUD", "NHA", "BDTHA" represent "Speed and Stopping Violations", "Right-of-Way and Traffic Control Violations", "Lane and Direction Violations", "Maneuvering and Signaling Errors", "General Unsafe Driving", "No Hazardous Action", "Both Drivers Took Hazardous Action", respectively.

*Table 2 Model Performance Evaluations.*

| Class (Support) | Llama 3.2 1B | Random Forest | Neural Network | XGBoost | CatBoost |
| --- | --- | --- | --- | --- | --- |
| SSV (35118) | **0.87 / 0.96 / 0.91** | 0.83 / 0.82 / 0.83 | 0.75 / 0.41 / 0.53 | 0.86 / 0.79 / 0.83 | 0.76 / 0.87 / 0.81 |
| RWTCV (32273) | **0.84** / 0.82 / **0.83** | 0.74 / 0.82 / 0.78 | 0.69 / 0.59 / 0.64 | 0.77 / 0.79 / 0.78 | 0.68 / **0.85** / 0.76 |
| LDV (8426) | 0.50 / **0.75 / 0.60** | 0.48 / 0.34 / 0.40 | 0.46 / 0.13 / 0.20 | **0.52** / 0.25 / 0.34 | 0.45 / 0.24 / 0.31 |
| MSE (5355) | 0.70 / **0.51 / 0.59** | 0.70 / 0.46 / 0.55 | 0.63 / 0.47 / 0.54 | **0.72** / 0.46 / 0.56 | **0.72** / 0.38 / 0.50 |
| GUD (2515) | **0.39** / 0.00 / 0.01 | 0.17 / 0.04 / **0.06** | 0.25 / 0.02 / 0.04 | 0.08 / **0.12** / 0.09 | 0.00 / 0.00 / 0.00 |
| NHA (2120) | **0.69** / 0.14 / **0.24** | 0.07 / 0.24 / 0.11 | 0.03 / **0.44** / 0.05 | 0.07 / 0.33 / 0.11 | 0.00 / 0.00 / 0.00 |
| BDTHA (1540) | **0.69** / 0.02 / 0.05 | 0.56 / 0.05 / 0.10 | 0.43 / **0.08 / 0.14** | 0.55 / 0.06 / 0.11 | 0.00 / 0.00 / 0.00 |
| **Accuracy** | **0.80** | 0.70 | 0.44 | 0.67 | 0.71 |
| **Macro Avg\*** (P / R / F1)\*\*\* | **0.67 / 0.46 / 0.46** | 0.51 / 0.40 / 0.40 | 0.46 / 0.31 / 0.30 | 0.51 / 0.40 / 0.40 | 0.37 / 0.33 / 0.34 |
| **Weighted Avg\*\*** (P / R / F1)\*\*\* | **0.79 / 0.80 / 0.77** | 0.72 / 0.70 / 0.70 | 0.66 / 0.44 / 0.51 | 0.74 / 0.67 / 0.69 | 0.65 / 0.71 / 0.67 |

*Note:*

*\*Macro Avg is the unweighted mean of the metric across all classes. It treats each class equally, regardless of how many instances it has, and is sensitive to performance in smaller classes.*

*\*\*Weighted Avg computes a class-weighted mean, where each class's metric is weighted by its sample size. This reflects the overall performance more in line with the class distribution.*

*\*\*\*P, R, F1 refer to Precision, Recall, and F1 score, respectively.*

## 4.2. Distribution-Shift Metrics for Hypothetical Risky Crash Scenarios

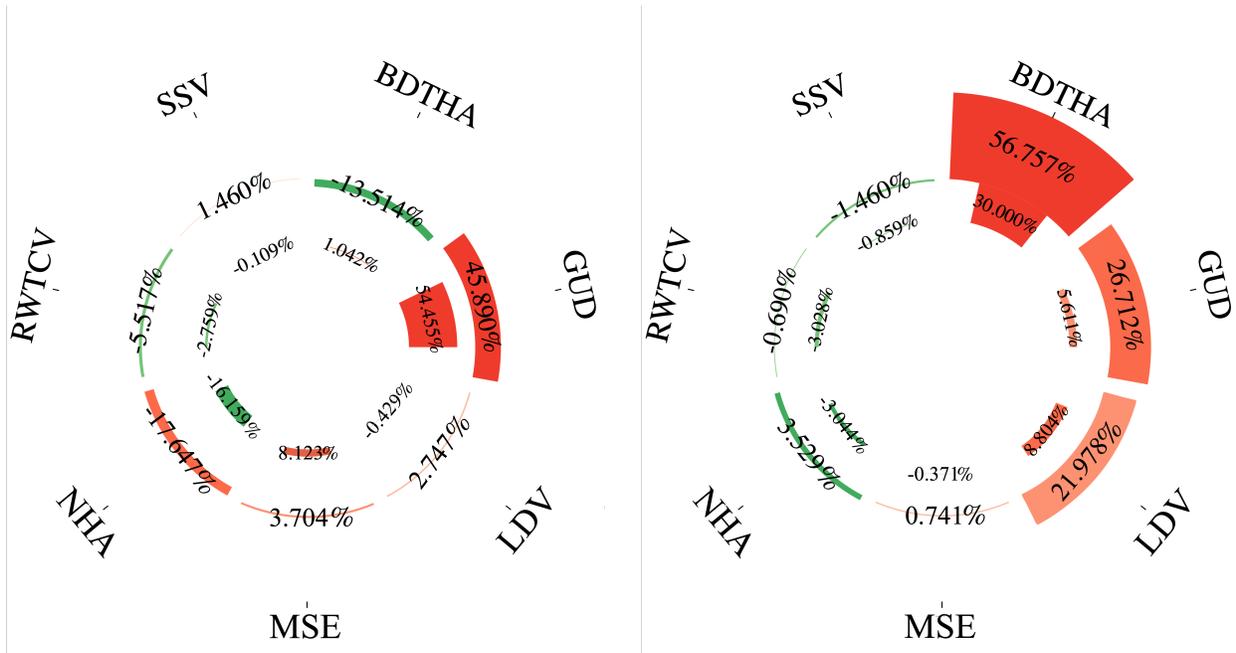

(a) Probability distribution shift between a baseline of 4% distraction and 50% distraction (single-driver distracted while the other is not).

(b) Probability distribution shift between a baseline of 4% distraction and 100% distraction (both-driver distracted).

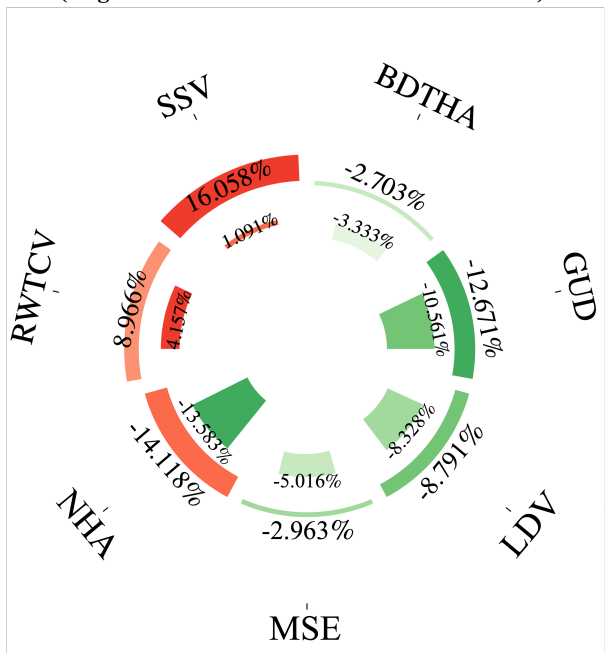

(c) Probability distribution shift between a baseline of 4% teen drivers and 100% distraction (both-driver are teen drivers).

*Note:*
*Inner Circle: IQR Change Proportion*
*Outer Circle: Median Change Proportion*

*All three simulated crash scenarios were conducted in the testing dataset, compared to the baseline (original testing dataset). Abbreviation explanations are listed below:*

- BDTHA – Both Drivers Took Hazardous Actions
- GUD – General Unsafe Driving
- LDV – Lane and Direction Violations
- MSE – Maneuvering and Signaling Errors
- NHA – No Hazardous Action
- RWTCV – Right-of-Way and Traffic Control Violations
- SSV – Speed and Stopping Violations

*Figure 5 Probability Distribution Shifts Between Baseline and Three Hypothetical Crash Scenarios.*

Figure 5 showed the relative shifts in our fine-tuned LLM's predicted-probability distribution, which were measured by median change proportion ($\Delta med$, Eq. 5) and IQR change proportion ($\Delta IQR$, Eq. 6), for each DHA class under three hypothetical crash scenarios, using the original testing set as the reference. It should be noted that this probability-based approach reflected relative tendencies rather than absolute risk because the predicted class probabilities necessarily sum to one after standardization, any upward shift in one class is accompanied by compensatory decreases elsewhere, so the observed increases and decreases should be interpreted as reallocations of the model's predicted probability rather than independent changes in DHA likelihood. A summary of observations for the three comparison results is provided below:

- **Figure 5a:** When each crash narrative in the testing set was synthetically modified so that one driver was relabeled "distracted" (originally accounted for 4%, now 50% among involved drivers) and the other "not distracted," our fine-tuned LLM reallocated probability mass most strongly toward General Unsafe Driving (GUD): the $\Delta med$ rose by +45.89 %, while the $\Delta IQR$ widened by +54.46 %. The simultaneous increase indicated not only a higher central tendency to label DHA as "GUD", but also markedly larger case-to-case variability, which was consistent with distraction introducing heterogeneous unsafe behaviors that the model captured under this class. Consistently, No Hazardous Action (NHA) showed a coherent contraction: $\Delta med = -17.65$ % and $\Delta IQR = -16.16$ %. The paired decline implied that as more distraction was engaged, the model became both less likely and less uncertain to assign the benign "NHA" label, reflecting an overall upward shift in perceived risk. Interestingly, Both Drivers Took Hazardous Action (BDTHA) exhibited the $\Delta med$ increased slightly (+1.04 %), yet its $\Delta IQR$ contracted sharply (–13.51 %). This pattern suggested the model did not systematically elevate or suppress the central probability of "BDTHA", but lower variation indicated the model was more confident about whichever probability it assigned, possibly because forcing only one driver to be distracted reduced the plausibility of both drivers simultaneously contributing HAs. The remaining classes, including Speed and Stopping Violations (SSV), Right-of-Way and Traffic Control Violations (RWTCV), Lane and Direction Violations (LDV), and Maneuvering and Signaling Errors (MSE) showed only minor shifts ($|\Delta| < 10\ \%$), which indicated relative stability of these HAs when distraction status was introduced single-driverally. Right-of-Way and Traffic-Control Violations (RWTCV) ticked a few non-intuitive downward percentage points ($\Delta med = -5.52\ \%$), which could be an artefact of probability-mass conservation once "GUD" captured the bulk of the gain as mentioned above. Since this decrease was small and counterintuitive, it should be interpreted cautiously as a residual balancing effect rather than evidence that distraction lowered perceived risk.

- **Figure 5b:** When both drivers in each crash narrative were synthetically relabeled "distracted" (originally accounted for 4%, now 100% among involved drivers), significant differences and similarities compared to scenario in Figure 5b were observed. First, our fine-tuned LLM reallocated probability most dramatically toward Both Drivers Took Hazardous Action (BDTHA). Its $\Delta med$ rose by +56.76 %, and the $\Delta IQR$ widened by +30.00 %, which indicated that "BDTHA" became the dominant explanation for crashes involving two distracted drivers and the model's predicted probability in that judgement is highly heterogeneous across cases. Second, General Unsafe Driving (GUD) still absorbed considerable mass ($\Delta med = +26.71\ \%$), yet its $\Delta IQR$ is relatively minor (+5.61 %) compared with the 54.46 % $\Delta IQR$ observed under single-driver distraction (see Figure 5a). The model therefore shifted many cases into "GUD" but did so with lower variances, which reflected reduced ambiguity once both drivers shared the same impairment. Interestingly, Lane and Direction Violations (LDV) emerged the third largest $\Delta med$ of +21.98 % while a relatively lower $\Delta IQR$ of +8.80 %. Relative to Figure 5a, the model reallocated probability from the "GUD" toward the more specific "LDV", implying that dual distraction sharpened cues associated with directional drift. Finally, Speed and Stopping Violations (SSV), Right-of-Way and Traffic-Control Violations (RWTCV), and Maneuvering and Signaling Errors (MSE) demonstrated consistent stability to the distraction allocation ($|\Delta| < 10\,\%$). Again, unexpected rises in No Hazardous Action (NHA) ($\Delta med = 3.53\,\%$), decreases in Speed and Stopping Violations (SSV) ($\Delta med = -1.46\,\%$), and Right-of-Way and Traffic-Control Violations (RWTCV) ($\Delta med = -0.69\,\%$) reflected the same residual balancing phenomenon described above and should not be interpreted as a genuine reduction in risk.

- **Figure 5c:** Substituting teenage drivers (aged 16 or 17, originally accounted for 4%, now 100% among involved drivers) into each crash narrative systematically shifted our fine-tuned LLM's predictions away from No Hazardous Action (NHA) ($\Delta med = -14.12\,\%$, $\Delta IQR = -13.58\,\%$), clearly reflecting narrowed probability variation in assigning DHAs to teen drivers. Two significant positive reallocations were observed: Speed and Stopping Violations (SSV) ($\Delta med = +16.06\,\%$, $\Delta IQR = +1.09\,\%$) and Right-of-Way and Traffic-Control Violations (RWTCV) ($\Delta med = +8.97\,\%$, $\Delta IQR = +4.16\,\%$). The relatively minor increases in their $\Delta IQR$s suggested that these DHAs became not only more prevalent in the teen-driver context but also more consistently predicted across cases. Conversely, the model reduced its probability allocations to General Unsafe Driving (GUD) ($\Delta med = -12.67\,\%$, $\Delta IQR = -10.56\,\%$), as well as Lane and Direction Violations (LDV) ($\Delta med = -8.79\,\%$, $\Delta IQR = -8.33\,\%$), Maneuvering and Signaling Errors (MSE) ($\Delta med = -2.96\,\%$, $\Delta IQR = -5.11\,\%$), and Both Drivers Took Hazardous Action (BDTHA) ($\Delta med = -2.70\,\%$, $\Delta IQR = -3.33\,\%$). These simultaneous reductions in both $\Delta med$ and $\Delta IQR$ indicated not only a lower central probability but also reduced variability, implying the model more consistently identified these classes as less tendential compared to the other three in the context. Given that the presence of residual balancing phenomenon, observed decreases in "GUD", "LDV", "MSE", and "BDTHA" should be interpreted primarily as reallocations of the model's predicted probability rather than mitigations in DHAs.

# 5. DISCUSSIONS

## 5.1. Fine-Tuning LLM for Advanced DHA Identification

This study proposed an analytical framework that leverages a fine-tuned LLM to advance the identification of DHAs in two-vehicle crashes. Specifically, the model utilized the Llama 3.2 1B architecture, selected for its computational efficiency, reformulating contextualized crash narratives as a causal next-token-generation task [15]. Through fine-tuning, the model adeptly captured linguistic structures and integrated extensive contextual information presented in reformulated crash narratives rather than traditional single crash record analysis, demonstrating substantial advantages over traditional classifiers and potential to be an automated alternative DHA coding technique. The framework addressed notable gaps identified in prior research by incorporating a comprehensive evaluation based on a large-scale dataset from MTCF. Unlike earlier NLP-based studies that focused on narrowly defined DHA identification tasks, this study thoroughly evaluated all valid DHA levels using a broad set of explanatory variables encompassing crash information, driver demographics, road characteristics, and environmental conditions [13, 16].

To ensure the model's practicality and interpretability, the original 14 officer-assigned DHA codes were systematically consolidated into seven behaviorally coherent classes, enabling clearer analysis and improved computational handling. This methodological choice improved computational efficiency and analytical clarity, particularly for rare categories such as "Speed Too Slow" (0.03% of records), which were grouped into broader class Speed and Stopping Violations (SSV). The fine-tuned LLM outperformed conventional classifiers, consistent with prior research showing that fine-tuning enhances LLM performance on domain-specific problems [18, 30]. Notably, the model excelled in predicting prevalent DHAs, such as Speed and Stopping Violations (SSV,

F1 = 0.91) and Right-of-Way and Traffic-Control Violations (RWTCV, F1 = 0.83), while still providing predictive utility for rarer classes (e.g., GUD, F1 = 0.01), albeit with reduced accuracy.

The LLM's superior performance offers significant practical implications for overcoming challenges in DHA coding, including inconsistency and labor-intensive manual processes. Traffic safety practitioners can leverage this framework as a decision-making tool, utilizing its accurate DHA prediction probabilities to standardize classification across large crash databases. For instance, the model's reliability in identifying SSV and RWTCV supports targeted safety interventions, such as enhanced enforcement or educational campaigns addressing these high-risk behaviors. However, the lower performance on rare classes necessitates human oversight to ensure robust interpretation, particularly in complex or less frequent scenarios. Moreover, when adapting this framework to other geographical contexts, practitioners may need to adjust DHA groupings or retain original codes to reflect local crash dynamics, underscoring the framework's flexibility and potential as an automated coding tool. By reformulating two-vehicle crash narratives rather than single-vehicle records, the framework introduced two additional meaningful DHA levels: "Both Drivers Took Hazardous Action" (BDTHA, 0.264% of narratives) and "No Hazardous Action" (NHA, 0.365% of narratives). These classes enhance the model's ability to capture shared-fault and no-fault scenarios, enriching its analytical scope.

## 5.2. Probabilistic Reasoning to Support Traffic Safety Interventions

Distraction and teen drivers are widely recognized as high-risk crash factors in the literature [31, 32, 33, 34, 35, 36, 37]. Here, we proposed a novel probabilistic reasoning approach that leveraged shifts in predicted DHA distributions generated by a fine-tuned LLM as decision-support tools for traffic safety interventions. By comparing each DHA's probability distribution shifts under the baseline and three hypothetical scenarios (single-driver distraction, both-driver distraction, and teen-driver substitution), our model offers insights into how the most significant DHA may redistribute under different crash conditions. Under the single-driver distraction situation where only one involved driver was distracted, the model predicted strong tendencies toward GUD and a corresponding decrease in NHA. This aligns with extensive empirical research as driver distraction is implicated in significant proportions of serious crashes, increases response delays, and leads to unsafe driving behaviors such as careless driving [31, 33, 35]. The diminished NHA probability further reflects the model's reduced uncertainty when an explicit distraction is present.

Under the both-driver distraction situation where both drivers were distracted, the LLM predicted a significant tendency towards BDTHA and LDV. Dual distraction is known to compound risk: when both drivers are cognitively impaired, opportunities for avoidance vanish, and both may exhibit erratic lane control or fail to adapt to intersections [36, 38]. When every driver was substituted with a teenager (ages 16–17), the model shifted probability toward SSV and RWTCV, which is consistent with observations from CDC data, which showed 35% of male and 18% of female teen drivers involved in fatal crashes were speeding [39]; previous studies [32, 34] similarly highlighted teen speeding's outsized role in crash causation. Though less frequently studied, distribution shifts generated by the LLM could serve as a valuable reference for understanding potential crash dynamics in underrepresented but risky scenarios. As such, these outputs may complement traditional empirical approaches by simulating plausible behavioral responses, thereby supporting proactive policy development and decision-making where direct observational data are limited or infeasible to collect.

# 6. CONCLUSION

In conclusion, this study develops and validates an analytical framework that leverages a fine-tuned LLM to identify DHA in two-vehicle crashes. A key finding is that the fine-tuned LLM significantly outperformed traditional machine learning models, demonstrating superior ability to capture nuanced contextual information embedded within crash narratives for accurate hazardous action prediction. Moreover, we introduce an innovative probabilistic reasoning approach that quantifies shifts in DHA predictions under various hypothetical high-risk scenarios, such as driver distraction and teen-driver involvement. The findings suggest that the developed framework can provide valuable insights to policymakers and traffic safety professionals, enabling a deeper understanding of how hazardous actions may shift under different circumstances and informing the design of more proactive safety interventions. By addressing the persistent challenges of effective DHA coding in existing crash datasets, our approach offers a scalable, automated alternative with the potential for broader applicability across diverse contexts. The comprehensive contextual modeling and validation procedures employed enhance both the reliability and generalizability of our analytical framework.

Nonetheless, several limitations should be acknowledged. First, relying exclusively on Michigan data may constrain the generalizability of the findings to other regions or national contexts. Second, recoding DHAs into broader categories could mask important nuances at the individual level, while excluding abnormal driver states—due to underreporting concerns—may omit relevant factors from analysis; this decision, however, was necessary to simplify modeling. Another limitation is the unexplored consistency of probability redistributions across different LLM architectures. Additionally, the inability to disaggregate DHAs at the individual driver level restricts insights into driver-specific behaviors. Finally, reformulating structured crash narratives, rather than utilizing original coder descriptions, may result in loss of valuable contextual information or pre-crash cues absent from tabulated databases.

Future research should focus on validating this framework using datasets from other states or national sources to boost its generalizability and exploring different LLM architectures for enhanced performance. Incorporating finer-grained DHA categories and abnormal driver behaviors could further increase both the model's accuracy and interpretability. Furthermore, leveraging richer narrative data—including original police reports—may enhance predictive capabilities and contextual depth. As LLM technology continues to advance, future models are expected to offer even stronger causal reasoning and reduced computational demands, further empowering data-driven traffic safety interventions.

## ACKNOWLEDGEMENT

The author would like to acknowledge the Toyota Collaborative Safety Research Center for sponsoring and supporting this project. Any opinions, findings, and conclusions, or recommendations expressed in this paper are those of the authors and do not necessarily reflect the views of Toyota.

## AUTHOR CONTRIBUTIONS

S.B. and H.G. contributed conceptualization, B.C.; K.Z. and G.X. contributed methodology; B.C. contributed software; B.C., G.X., and K.Z. contributed validation; B.C. contributed formal analysis; B.C. contributed data curation; B.C., G.X., Z.W., K.Z., H.G., A. A., Z.S., Z.H., and S.B. contributed writing original draft preparation; B.C., G.X., Z.W., K.Z., H.G., A. A., Z.S., Z.H., and S.B. contributed writing review and editing; B.C. and S.B. contributed visualization; S.B. and H.G. contributed supervision. All authors have read and agreed to the published version of the manuscript.